\documentclass[letterpaper, 10 pt, conference]{ieeeconf}  

\IEEEoverridecommandlockouts                              

\usepackage{graphics} 
\usepackage{epsfig} 
\usepackage{amsmath} 
\usepackage{amssymb}  
\usepackage{multirow}
\usepackage[switch]{lineno}
\usepackage{nameref}
\usepackage{hyperref}
\usepackage{todonotes}
\usepackage{bm}
\usepackage{dblfloatfix}

\title{\LARGE \bf
  Proactive Interaction Framework for Intelligent Social Receptionist Robots
}

\author{Yang Xue$^{1}$, Fan Wang$^{1}$, Hao Tian$^{2}$, Min Zhao$^{3}$, Jiangyong Li$^{1}$, Haiqing Pan$^{3}$ and Yueqiang Dong$^{1}$
\thanks{$^{1}$Yang Xue, Fan Wang, Jiangyong Li and Yueqiang Dong are from Baidu Natural Language Processing Department (e-mail: {\{xueyang02,wang.fan,lijiangyong01,dongyueqiang\}@baidu.com};)}%
\thanks{$^{2}$Hao Tian is from Baidu Research, (e-mail: {tianhao@baidu.com})}%
\thanks{$^{3}$Min Zhao and Haiqing Pan are from Baidu AI Interaction Design Lab, (e-mail:  {\{zhaomin04,panhaiqing\}@baidu.com})}
}

\begin{document}

\maketitle
\thispagestyle{empty}
\pagestyle{empty}

\begin{abstract}
Proactive human-robot interaction (HRI) allows the receptionist robots to actively greet people and offer services based on vision, which has been found to improve acceptability and customer satisfaction.
Existing approaches are either based on multi-stage decision processes or based on end-to-end decision models. 
However, the rule-based approaches require sedulous expert efforts and only handle minimal pre-defined scenarios. On the other hand, existing works with end-to-end models are limited to very general greetings or few behavior patterns (typically less than 10). 
To address those challenges, we propose a new end-to-end framework, the \textbf{T}rans\textbf{F}ormer with \textbf{V}isual \textbf{T}okens for \textbf{H}uman-\textbf{R}obot \textbf{I}nteraction (TFVT-HRI)\footnote{Source code repository: \url{https://github.com/PaddlePaddle/PaddleRobotics/tree/ICRA21/HRI/TFVT_HRI}.}. The proposed framework extracts visual tokens of relative objects from an RGB camera first.  To ensure the correct interpretation of the scenario, a transformer decision model is then employed to process the visual tokens, which is augmented with the temporal and spatial information. It predicts the appropriate action to take in each scenario and identifies the right target. Our data is collected from an in-service receptionist robot in an office building, which is then annotated by experts for appropriate proactive behavior. The action set includes 1000+ diverse patterns by combining language, emoji expression, and body motions.
We compare our model with other SOTA end-to-end models on both offline test sets and online user experiments in realistic office building environments to validate this framework. It is demonstrated that the decision model achieves SOTA performance in action triggering and selection, resulting in more humanness and intelligence when compared with the previous reactive reception policies.
\end{abstract}

\section{INTRODUCTION}\label{sec:intro}

Receptionist robots work in public areas such as lobbies and shopping malls, helping to guide visitors, post instructions, and answer questions~\cite{nisimura2002aska,hashimoto2007realization,niculescu2011influence}.
As most existing receptionist robots can only passively answer users' calls, new visitors may have little motivation to initiate an interaction with the robot in many cases.  This can be attributed to the lack of knowledge of the robot or low-expectations of its capability.
While proactive reception robots can try to initiate interactions themselves, there are also risks of arousing antipathy in inappropriate proactive behavior cases.

\begin{figure}
  \centering
  \includegraphics[width=0.6\columnwidth]{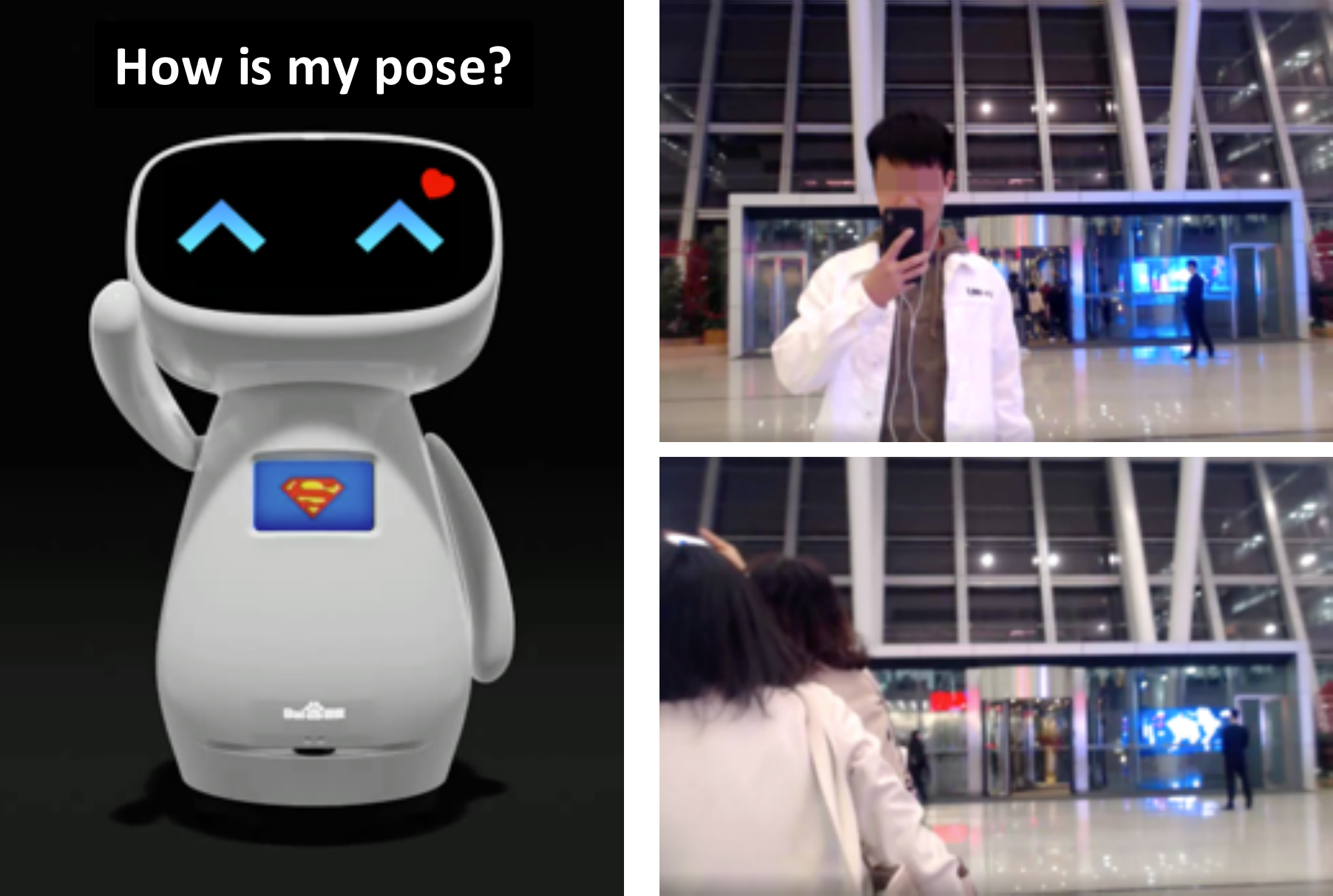}
  \caption[Caption without FN]{\textbf{Left}: An example of proactive multi-modal behavior of
    the \textit{Xiaodu} 
    robot for users who are taking a photo ahead of it.
    Specifically, the robot is making a superman pose and saying
    ``How is my pose''. Here we use the avatar for better visualization.
    Check Fig.~\ref{fig:xiaodu} for more information on \textit{Xiaodu} robot.
    \textbf{Right}: Two different ``photo-taking'' scenarios in the served lobby.}\vspace{-0.2in}
  \label{fig:diverse_photo_taking}
\end{figure}

To make proactive interaction more human-like and socially acceptable, it is desirable to correctly understand user intention.
Previous works are based on empirical social rules and require the specification of scenarios at first, such as ``A human is passing by,'' or ``A human is waving a hand towards the robot.''
To identify those scenarios, sophisticated sensor settings~\cite{kato2015may,bergstrom2008modeling,ozaki2018decision} are commonly required.
Further, there are ``micro signals'' that are too complicated to be categorized.
For instance, it is common for visitors to take a photo of the reception robot alone or take a group photo with the robot, shown in Fig.~\ref{fig:diverse_photo_taking}.

\addtocounter{footnote}{-1}


While prior researchers also propose to use an end-to-end framework in generating proactive greetings \cite{ozaki2019can,tuyen2018emotional,doering2019modeling}, their works suffer from the following limitations:
First, the robot behaviors are restricted to very few types, typically less than 10, most of which are general words such as ``Hello''\cite{zhao2019stepped,heenan2014designing}, ``Excuse me'', ``May I Help You''~\cite{kato2015may}.
Although those words can ``break the ice'' to some extent, they provide little information and show no understanding of the scenario.
Second, it is desirable to reason over time instead of depending on one single frame for proactive HRI.
For instance, a visitor may hesitate and look around for a long time, which is a vital sign of seeking help.
Third, the robot may be required to target a specific person or a particular group of persons instead of the scenario, especially when multiple persons are in sight, while most end-to-end decision-making is scenario-based.

To address aforementioned challenges, we propose a new end-to-end framework namely \textbf{T}rans\textbf{F}ormer with \textbf{V}isual \textbf{T}okens for Human-Robot Interaction (TFVT-HRI).
It has the following unique features:
First, to reason over time while targeting specific persons, we use a \emph{visual token extractor} to turn image-level signals into object-level tokens.
Second, we use consecutive frames of a video clip, along with the temporal and spatial information of each token. Third, we use a transformer model to process the tokens. The attention over visual tokens enhances information exchange among different objects and different times to better interpret the scenario.
Fourth, to improve the generalization ability of our model, we employ natural language pretraining~\cite{sun2019ernie} as well as image pretraining in the representation layers.
To train the decision model, we collect a rich dataset from in-service reception robots in office-building lobbies, where our staff imitate the daily visitors by implying different interaction intentions.
We then invite experts to review those videos and label appropriate proactive behaviors.
We collect over 1000+ multi-modal actions composed of language, emoji expression, and body movements.
The action set includes not only simple greetings such as ``Good morning, miss'', but also scenario-specific reactions such as ``How is my pose'' (Fig.~\ref{fig:diverse_photo_taking}), ``Are you interested in playing a game with me?'' and ``Are you looking for some places?''
It not only permits more human-like interactions but is also able to guide the visitors to further communications in depth.

To validate the proposed method, we first compare TFVT-HRI with the state-of-the-art action recognition method, fine-tuned R(2+1)D~\cite{ghadiyaram2019LargeScaleWP} model on the test set of our collected data, which shows that TFVT-HRI pushes on to frontier in this task.
We then conduct a user experiment on the \textit{Xiaodu} robot (Fig.~\ref{fig:xiaodu}), a receptionist robot serving in office buildings, museums, etc.
It is shown that TFVT-HRI can generate human-like behaviors showing a deep understanding of the scenario and achieves a substantially higher score in overall comfort, naturalness, friendliness, and intelligence than prior wake word-based HRI system of the \textit{Xiaodu} robot.

\section{RELATED WORK}
\label{sec:related_work}

\textbf{Social Rule based Systems}.
A large bunch of the previous works is based on social rules to initiate interaction with a human.
Some of them make a straightforward classification of two classes: whether the user has the intention of interaction~\cite{kato2015may,ozaki2018decision}.
The robot behavior includes a single pattern greeting only.
Bergstrom et al.~\cite{bergstrom2008modeling} utilize a laser range finder to collect the motion of the visitors around the robot, by which the visitors are classified into four groups depending on their trajectories, revealing the different level of interest in the robot. The robot behaviors are then specified based on the categorization.
Heenan et al.~\cite{heenan2014designing} derive action set from Kendon's theory of conducting interaction~\cite{kendon1990conducting}.
The classification is based on the user's distance and head orientation, and the robot action is derived from Hall's proxemics theory~\cite{hall1966hidden}, which is also suggested by Zhao et al.~\cite{zhao2019stepped}.
Further, Zhao et al.~\cite{zhao2019stepped} extend proactive HRI to progressive initiations including visual observation and eye contact.
Those approaches inevitably require the pre-definition of scenarios.

\textbf{Imitation Learning for Human-Robot Interaction}.
Imitation learning is frequently employed to enable robots to learn complex human skills.
For HRI, it has been used to instruct a robot to express emotional body language~\cite{tuyen2018emotional}, and teach a robot to play the role of a travel agent~\cite{doering2019modeling} or a shopkeeper~\cite{liu2018learningPB}, showing that learning from human demonstrations can enable a robot to learn communication skills.
However, imitation learning has not been widely used in the proactive HRI task yet.

\textbf{Human Intention and Video Action Recognition}.
To understand the human intention, besides tracking the motion, speed, etc. of human pedestrians with laser and other sensors~\cite{kato2015may,ozaki2018decision}, it is also possible to use video clips or images.
The recently proposed pre-trained R(2+1)D model~\cite{ghadiyaram2019LargeScaleWP} aims to the elaborate classification of human action, which is a competitive approach in proactive action generation.
However, to generate correct behaviors we still need an exhaustive specification of reacting rules after classification.
Further, in the case of multiple visitors, designing rules can be too complicated.
We provide a detailed comparison of our method with action recognition methods in our experiments.


\textbf{Metrics to Evaluate Interaction Initiation}.
In previous studies, proactive HRI is mainly evaluated by system performance and users’ subjective evaluations.
For system performances, Liao et al.~\cite{liao2016what} and Rashed~\cite{rashed2016} use the success rate of initiation of interaction; Shi et al.~\cite{shi2015behavior} evaluate the recognition accuracy of participants' state.
For users' subjective evaluations, Zhao et al.~\cite{zhao2019stepped} and Ozaki et al.~\cite{ozaki2018decision} utilize questionnaire to study users' experiences. Liao et al.~\cite{liao2016what} consider the feedbacks on satisfaction, likability, perceived interruption, perceived hedge, and social-agent orientation. Bergstrom et al.~\cite{bergstrom2008modeling} and Shi et al.~\cite{shi2015behavior} mainly focus on the naturalness or appropriateness of the robot behavior.
In this work, we will evaluate our framework from both offline datasets and online questionnaires.

\section{METHODOLOGY}
\label{sec:method}

\begin{figure*}
  \centering
  \vspace{0.08in}
  \includegraphics[width=2\columnwidth]{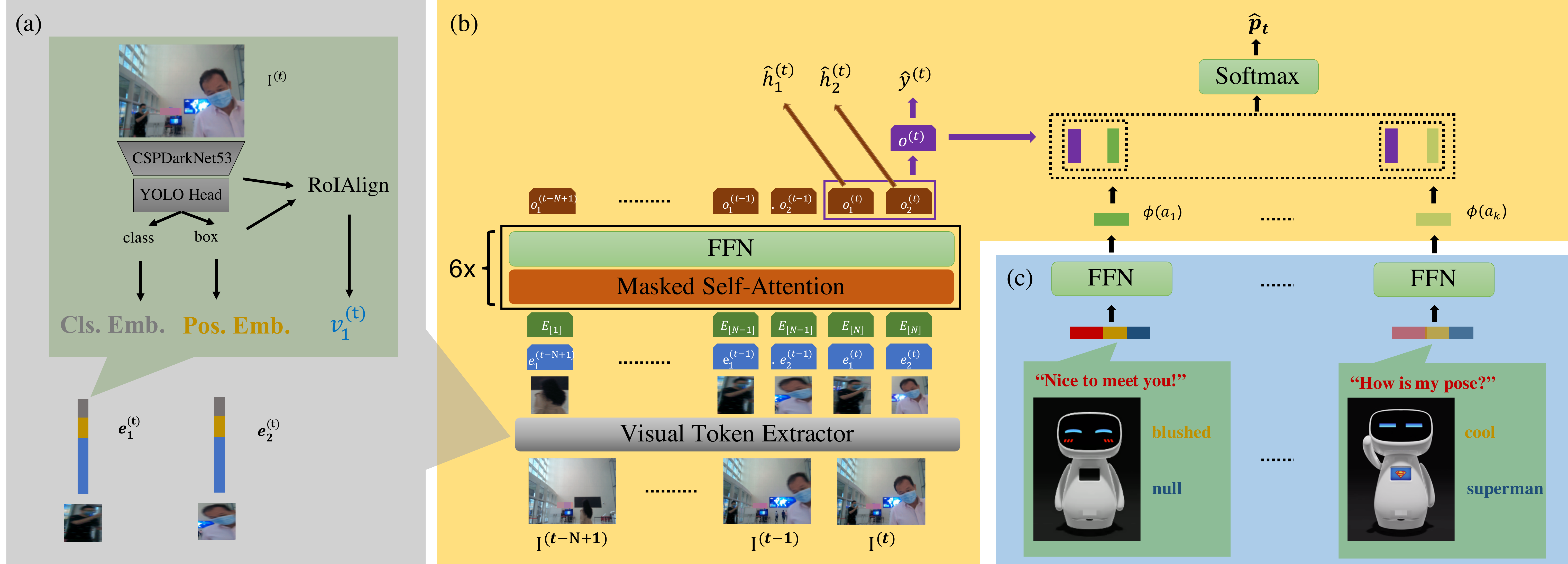}
  \caption{Illustration of the TFVT-HRI framework, which is composed of three modules:
    (a) Visual token extractor that extracts visual tokens of objects in one frame.
    (b) Transformer-based decision model which encodes video clip to predict proactive behaviors
    (c) Multi-modal action encoder that encodes the natural language, expression, and body motion to form a representation vector.}\vspace{-0.2in}
  \label{fig:framework}
\end{figure*}

The proposed framework takes sequential RGB frames as the inputs and decides whether to initiate an interaction, which target to interact, and what the action to do.
The overview of the framework is illustrated in Fig.~\ref{fig:framework}, which is composed of three modules: visual token extractor (Fig.~\ref{fig:framework}(a)); multi-modal action encoder (Fig.~\ref{fig:framework}(c)); and transformer-based decision model (Fig.~\ref{fig:framework}(b)).

\subsection{Visual Token Extractor}
\label{sec:vt-extractor}
We denote the input image flow as $I^{(1)}, I^{(2)}, ..., I^{(t)}$, where we use superscript $(t)$ to represent the $t$-th frame. In this module, we first extract the region of interests by YOLOv4 \cite{bochkovskiy2020YOLOv4OS}.
YOLOv4 is a powerful object-detection model that detects a wide range of classes of objects. In our case, we are mainly interested in 6 categories that may be connected to human identity and action recognition in our scenarios, including ``person'', ``backpack'', ``handbag'', ``suitcase'', ``tie'', and ``cell phone'', while other categories are neglected.

To reduce the computation cost, we re-use the output of the backbone of YOLOv4, the CSPDarkNet53 \cite{wang2020cspnet}, as the feature extractor. As the bounding boxes are in different sizes, we apply RoIAlign pooling to normalize the sizes of feature maps, inspired by the MaskRCNN \cite{he2017MaskR}, as well as global average pooling (GAP) to reduce the feature representation dimension.
Through this process we are able to represent each pedestrian $i$ in the image $I^{(t)}$ with a 512-dimensional feature vector $v^{(t)}_i$. We also add the embedding vector of 2D position information \cite{carion2020EndtoEndOD} and object classification ID, which gives the final visual token of ($\oplus$ denotes ``concatenation" and $\textit{EMB}$ represents ``embedding vector'')
\begin{equation}\label{eq:visual_token}
  e^{(t)}_i = v^{(t)}_i \oplus \textit{EMB}(\textit{POS}(i,t)) \oplus \textit{EMB}(\textit{CLASS}(i,t)).
\end{equation}

\subsection{Multi-modal Action Encoder}
\label{sec:act-repr}
The action is denoted as a combination of an utterance ($u$), an emoji facial expression ($f$), and a body motion ($m$). We specify a static set of distinct actions $a_k=(u_k, f_k, m_k), k \in [1,K]$ by collecting the actions from the annotated dataset.
To better represent the action in semantic space, we utilize an open-source natural language pre-training model ERNIE~\cite{sun2019ernie} to process each utterance into a vector. Then facial expression and the motion are limited to several pre-defined patterns, such as ``smiling'' and ``hand-shaking'' (Fig.~\ref{fig:xiaodu}). For each type of body motion and expression, we use an embedding vector. The representation of three modalities are concatenated and further processed by an additional feed-forward neural layer (FFN) to yield the representation for $k$-th action, denoted by

\vspace{-0.13in}
\begin{equation}\label{eq:action representation}
  \phi(a_k) = \textit{FFN}(\textit{ERNIE}(u_k) \oplus \textit{EMB}(f_k) \oplus \textit{EMB}(m_k)).
\end{equation}

\subsection{Transformer-based Decision Model}
\label{sec:transformer-hri}
To enable the decision model to reason over time, normally, we could try to track each individual and encode the trajectory. Unfortunately, tracking one individual requires registration of visual tokens across different frames, which can introduce additional error to the decision. To this end, instead of tracking each individual, we present a transformer decision model. Transformer is widely used in natural language processing~\cite{vaswani2017attn} by allowing the tokens to fully interact with each other through self-attention. In our case, it is not only able to capture the trajectory of each pedestrian by attending over time, but also possible to capture the interaction between the target and other individuals, e.g., talking with another person, holding a cell phone, carrying a suitcase. Also, to preserve consistency for training and inference, for each visual token $e^{(t)}_i$ that we consider, we allow it to attend to $e^{(t')}_i$ only when $t' \leq t$, which implies masked self-attention in transformer blocks (MTRN for short). In our case, we use 6 transformer blocks.

For each frame $I^{(t)}$ received, we use visual token extractor to acquire a list of visual tokens $e^{(t)}_1, e^{(t)}_2, ...$, from which we consider the top-$M$ ($M=20$ in the experiments) visual tokens. The priority of the selection goes as follows: the ``person'' class is selected in the first place; the larger the bounding box, the higher priority; in case there are less than $M$ visual tokens, we add padding terms. In every time step, we use the latest $N$ frames as the input, which gives $M \times N$ tokens in one sequence. For each token, we further add an embedding vector of the relative frame ID $\textit{E}_{[i]}$. The encoding process can be represented by

\vspace{-0.2in}
\begin{align}
\label{eq:masked_self_attn}
o^{(t-N+1)}_1,& ..., o^{(t-N+1)}_M, ..., o^{(t)}_M = \textit{MTRN}(\nonumber\\
&e^{(t-N+1)}_1 \oplus \textit{E}_{[1]}, ..., e^{(t-N+1)}_M \oplus \textit{E}_{[1]}, \nonumber\\
&e^{(t-N+2)}_1 \oplus \textit{E}_{[2]}, ..., e^{(t-N+2)}_M \oplus \textit{E}_{[2]}, \nonumber\\
&,..., \nonumber\\
&e^{(t)}_1 \oplus \textit{E}_{[N]}, ..., e^{(t)}_M \oplus \textit{E}_{[N]})
\end{align}

Starting from Eq.~\ref{eq:masked_self_attn}, the prediction of the decision model is three-fold: 1. decide whether to initiate an interaction, which depend on $\hat{y}^{(t)}$; 2. predict the target to interact, which is dependent on $\hat{h}^{(t)}_i$;  3. select the action to be taken, which is dependent on $\hat{p}^{(t)}(a_k)$. We first take the max-pooling of all the positions in the last frame to represent the scenario, which is defined as

\vspace{-0.1in}
\begin{equation}
\label{eq:frame_pool}
o^{(t)} = \textit{MaxPooling}(o^{(t)}_1, o^{(t)}_2, \ldots, o^{(t)}_M),
\end{equation}
we then specify the three-fold output of the decision model as follows:
\begin{align}
\label{eq:output_decision_model}
\hat{y}^{(t)} =& \sigma(W_y \cdot o^{(t)} + b_y) \\
\hat{h}^{(t)}_i =& \sigma(W_h \cdot o^{(t)}_i + b_h)) \\
\hat{p}^{(t)}(a_1),...,p^{(t)}(a_k) =& Softmax(\phi(a_1) \cdot o^{(t)}, ..., \nonumber\\
&\phi(a_k) \cdot o^{(t)})
\end{align}

Correspondingly, the annotation in training data include three parts: interaction trigger $y^{(t)}=0/1$, where $y^{(t)}=1$ indicate a proper opportunity to initiate the interaction; interaction target indicator $h_i^{(t)} = 0/1$, where $h_i^{(t)}=1$ implies that the visual token is one of the target to be interacted; the action selection $\bm{I}^{(t)}_a$ which is a one-hot vector of length $K$. We use cross-entropy loss $\mathcal{L}_{\textit{ce}}$ for each of them, giving the loss function of Eq.~\ref{eq:loss}.
\begin{align}
\label{eq:loss}
\mathcal{L} =& \sum_t \mathcal{L}_{\textit{ce}}(y^{(t)}, \hat{y}^{(t)}) + \sum_t y^{(t)} \cdot \mathcal{L}_{\textit{ce}}(\bm{I}^{(t)}_a, \bm{\hat{p}}^{(t)}) \nonumber\\
  &+ \sum_t y^{(t)} \sum_i \mathcal{L}_{\textit{ce}}(h^{(t)}_i,  \hat{h}^{(t)}_i)
\end{align}

For inference, we use weighted sampling on $\bm{\hat{p}}^{(t)}$ for action selection; for target filtering, we first remove the visual tokens of classes other than ``person'', then we use a threshold $H$, such that $h^{(t)}_i > H$ indicate a valid target. At executing the actions, the robot turns towards the centroid of all the valid targets. We argue that turning to the interaction target is essential to the user experience.

\section{EXPERIMENTAL RESULTS}

\subsection{Data Collection}

\begin{figure}
  \centering
  \vspace{0.08in}
  \includegraphics[width=0.8\columnwidth]{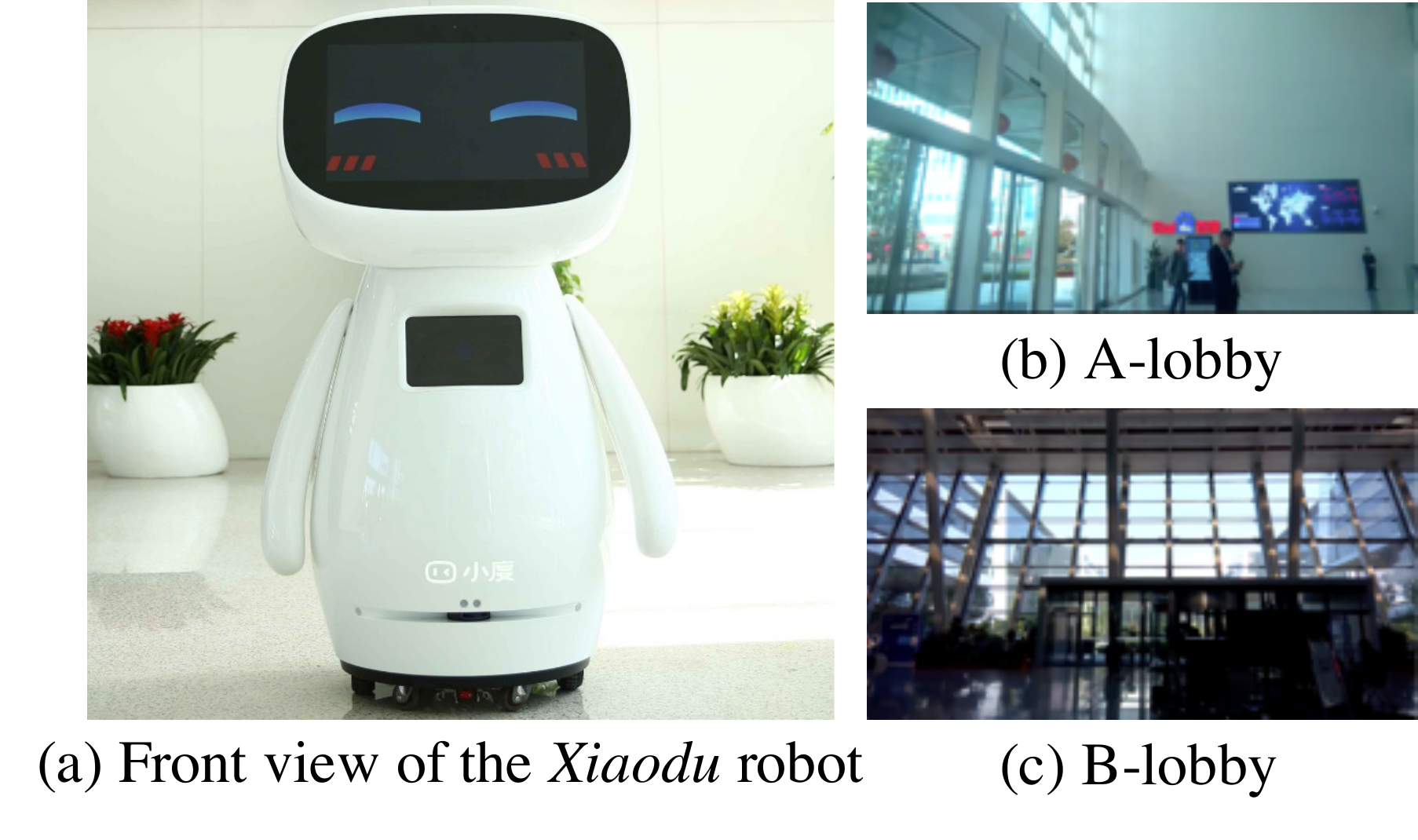}
  \vspace{-0.1in}
  \caption{(a) The \textit{Xiaodu} robot, a receptionist robot that we used to conduct all the experiments. An RGB camera is on its head to record live video that serves as input to the model. Execution of the multi-modal action includes three modalities: its stereo speaker can play synthesized TTS given the utterance; a LED screen on the face can display over 30 emojis; for motions, it can perform hand-shaking, hand-waving, hug, moving, etc.
  (b) The A-lobby from the perspective of the robot.
  (c) The B-lobby from the perspective of the robot.}\vspace{-0.2in}
  \label{fig:xiaodu}
\end{figure}

Hours-long videos were collected from two working \textit{Xiaodu} robots (Fig.~\ref{fig:xiaodu}) in the office lobbies, and then they were annotated and processed into 5-second-long video clips as the training and test set.
The office lobbies for our data collection are significantly different in the light condition.
As shown in Fig.~\ref{fig:xiaodu} (b) and (c), the B-lobby has a worse light condition than the A-lobby because of the backlighting environment. This makes the vision-based approaches more challenging in the B-lobby.
After collecting hours-long videos from these two lobbies, we used the following method to preprocess, annotate, and post-process them:
\begin{enumerate}
  \item The raw videos were preprocessed by a multiple object tracking model~\cite{wojke2017SimpleOA}, which marks the bounding boxes of tracked persons with a unique ID for each. The experts label suitable targets for initiating interactions with track IDs.
  \item Experts watched the preprocessed videos and selected the suitable timestamps for initiating interactions. For each selected timestamp, the expert annotated the tracking IDs of targeting interaction objects, chose the facial expression and the body motion from a list that the \textit{Xiaodu} robot permits, and filled the textbox with a proper utterance for greeting.
  \item Annotated videos were post-processed into 5-second-long video clips in which positive examples were from experts' annotations, and negative examples were randomly sampled from segments without annotations then filtered out in-interaction segments.
\end{enumerate}

Statistically, for the training set, we collected and labeled 3900 positive cases (i.e., trigger label is 1) with 1000+ unique multi-modal actions (triplet of utterance to speak, emoji to display, and body action to execute), in which 1720 cases are from B-lobby, and 2180 cases are from A-lobby.
For the test set, from 12-hour videos, we collected and labeled 74 positive cases in A-lobby and 91 positive cases in B-lobby.
However, when evaluating the performance on the test set, we used the full 135k+ negative cases instead of sampling to verify the model sufficiently.

\subsection{Offline Evaluation}

\begin{table*}[!b]
\centering
\caption{Performance of model variants and results of ablation study.}
\label{tab:perf_comp}
\begin{tabular}{ |cc|c|c|c|c|c|c|c|c|c|c| }
\hline
\multicolumn{2}{|c|}{\multirow{2}{*}{\textbf{Methods}}} & \multicolumn{5}{|c|}{\textbf{A-Lobby}} & \multicolumn{5}{|c|}{\textbf{B-Lobby}} \\ \cline{3-12}
 &  & \textbf{Precision} & \textbf{Recall} & \textbf{AP} & \textbf{AR} & \textbf{F1} & \textbf{Precision} & \textbf{Recall} & \textbf{AP} & \textbf{AR} & \textbf{F1} \\
\hline
R(2+1)D+ig65m & & 0.383 & 0.837 & - & - & 0.526 & 0.267 & 0.859 & - & - & 0.407 \\
\hline\hline
\multirow{3}{*}{TFVT-HRI} & Trigger-Only & 0.905 & 0.851 & 0.718 & \textbf{0.927} & 0.877 & 0.583 & 0.778 & 0.418 & \textbf{0.949} & \textbf{0.667} \\
& Actor-Only & 0.894 & 0.868 & - & - & \textbf{0.881} & 0.485 & 0.876 & - & - & 0.624 \\
& Trigger-Actor & 0.905 & 0.851 & - & - & 0.877 & 0.585 & 0.775 & - & - & \textbf{0.667} \\
\hline\hline
\multirow{3}{*}{TFVT-HRI w/o $v^{(t)}_i$} & Trigger-Only & 0.717 & 0.930 & 0.713 & 0.857 & 0.810 & 0.382 & 0.989 & \textbf{0.434} & 0.894 & 0.551 \\
& Actor-Only & 1.0 & 0.121 & - & - & 0.216 & 0.800 & 0.148 & - & - & 0.250 \\
& Trigger-Actor & 1.0 & 0.061 & - & - & 0.114 & 1.0 & 0.041 & - & - & 0.078 \\
\hline
\multirow{3}{*}{TFVT-HRI w/o Pos. Emb.} & Trigger-Only & 0.782 & 0.910 & 0.755 & 0.893 & 0.841 & 0.420 & 0.941 & 0.399 & 0.929 & 0.581 \\
& Actor-Only & 0.831 & 0.790 & - & - & 0.810 & 0.424 & 0.911 & - & - & 0.579 \\
& Trigger-Actor & 0.864 & 0.691 & - & - & 0.768 & 0.405 & 0.757 & - & - & 0.527 \\
\hline
\multirow{3}{*}{TFVT-HRI w/o Cls. Emb.} & Trigger-Only & 0.855 & 0.810 & \textbf{0.786} & 0.738 & 0.832 & 0.411 & 0.966 & 0.420 & 0.861 & 0.576 \\
& Actor-Only & 0.700 & 0.212 & - & - & 0.326 & 0.314 & 0.800 & - & - & 0.451 \\
& Trigger-Actor & 0.714 & 0.152 & - & - & 0.251 & 0.337 & 0.648 & - & - & 0.443 \\
\hline
\end{tabular}
\end{table*}

To evaluate the proactive HRI framework, it is desirable to verify whether the system predicts the correct timestamp to initiate interaction and whether the action selection is proper. However, although each positive case in the test set has a labeled action, we found that multiple actions can be suitable for one scenario. Therefore, evaluating the action selection is extremely challenging. In this part, we evaluated the capability of predicting proper triggering time only.
The user experience research in the next part will evaluate whether the selected action is proper.

We used the state-of-the-art model called R(2+1)D~\cite{ghadiyaram2019LargeScaleWP} in a similar video action recognition task as our baseline.
The R(2+1)D model was pre-trained on a large video dataset from social media. We fine-tuned it on our dataset in an end-to-end manner to predict the multi-modal action ID.
To enable a fair comparison between our Transformer setting and R(2+1)D, we add a ``NULL'' action in the action set, such that selecting the null action is equal to not triggering an initiation. For R(2+1)D, the output is a $(K+1)$-way classification on the action space. In this way, the R(2+1)D can predict the triggering time without predicting $\hat{y}^{(t)}$.
For our model, we also add a ``NULL'' action in the action set, this leads to three different inference modes:
\begin{itemize}
    \item \textbf{Trigger-Only} By removing the null action, We rely only on $\hat{y}^{(t)}$ to decide whether to initiate interaction.
    \item \textbf{Actor-Only} By setting $\hat{y}^{(t)}=1$ identically, we rely only on  $\bm{\hat{p}}^{(t)}$ to select the null action to serve as the trigger.
    \item \textbf{Trigger-Actor} Initiation is only triggered when we both have $\hat{y}^{(t)} > H$ and avoid selecting the null action.
\end{itemize}

To find out the contribution of different elements in the visual token, we designed ablation studies by removing the RoIAlign pooling feature ($v^{(t)}_i$), the position embedding, and the classification embedding in turn. We reported the performance of successful proactive interaction (proactively and appropriately interact with the user) in terms of precision and recall.
The performance of the trigger is related to the threshold, so we include two more metrics: average precision (AP) and average recall (AR) for better evaluation.
Furthermore, we use the F1 score to demonstrate overall performance. All the results are shown in Table~\ref{tab:perf_comp}.
Note that we select the threshold with the maximum F1 score to report the precision and recall.
In the Trigger-Actor mode, we also use this threshold for the trigger.
As we can see, the results show that the proposed transformer model outperforms the state-of-the-art end-to-end R(2+1)D model.
Besides, the ablation study shows that the most important factor is the information from the RoIAlign pooling feature ($v^{(t)}_i$). Without this information, the framework suffers the largest F1 score drop.

\begin{table*}[!b]
\centering
\caption{Descriptive statistic results of user questionnaires.}
\label{tab:user_reports}
\begin{tabular}{|c|c|c|c|c|c|c|c|c|c|c|c|c|c|}
\hline
\multirow{3}{*}{\textbf{Methods}} & \multirow{3}{*}{\textbf{Particiants}} & \multicolumn{4}{c|}{\textbf{Emotion}}                                & \multicolumn{8}{c|}{\textbf{Attitudes}}                                                                                                 \\ \cline{3-14}
                         &                              & \multicolumn{2}{c|}{\textbf{Valence}} & \multicolumn{2}{c|}{\textbf{Arousal}} & \multicolumn{2}{c|}{\textbf{Comfort}} & \multicolumn{2}{c|}{\textbf{Naturalness}} & \multicolumn{2}{c|}{\textbf{Friendliness}} & \multicolumn{2}{c|}{\textbf{Intelligence}} \\ \cline{3-14}
                         &                              & M             & SD           & M             & SD           & M             & SD           & M               & SD             & M               & SD              & M               & SD              \\ \hline
TFVT-HRI          & 15                           & 6.20          & 1.93         & 6.27          & 2.15         & 5.20          & 1.42         & 4.87            & 1.41           & 6.07            & 1.22            & 4.53            & 1.73            \\ \hline
Reactive HRI & 15                           & 5.27          & 2.25         & 4.72          & 2.25         & 4.13          & 1.30         & 3.73            & 1.22           & 4.67            & 1.59            & 3.33            & 1.18            \\ \hline
\end{tabular}
\end{table*}

\subsection{User Experience Research}


Because the data were collected under a passive HRI system, the distribution of the test set was different from that of the online experiments.
To figure out whether the proposed framework performs well on a real robot, we implement this user experience research on a real robot in the A and B lobbies, comparing the existing wake word-based interaction system of the \textit{Xiaodu} robot.
On the Jetson AGX Xavier, the proposed TFVT-HRI model runs at 6.25fps, whereas the R(2+1)D model runs at 1.89fps.
Since the R(2+1)D model suffers from 3x computation latency, and its performance is worse than the proposed framework, we do not deploy it on the real robot for this user experience research.

\textbf{Participants}. We recruited 30 new employees who have never interacted with the \textit{Xiaodu} robot, but 26 of them used other similar AI devices (they share the same wake word ``X'' to start an interaction).
Among the participants, there are 16 male and 14 female; their ages range from 21 to 35 years old (M=27.1, SD=3.84).
All participants reported normal or corrected to normal vision and normal hearing.
All participants volunteered to participate in the study and agreed to make audio and video recordings of the research process.
At the end of the study, all participants were given appropriate compensation.

\textbf{Design}. The experiment adopted a between-subjects design.
Participants were randomly assigned to one of the two groups, the experimental group and the control group.
In the experimental group, the robot was deployed with the proposed framework and would proactively initiate interaction.
In the control group, the robot was waiting reactively for the wake word.

The dependent variables included the objective factor (success rate) and subjective factors (emotions, attitudes).
For emotions, we utilize the Self-Assessment Manikin (SAM) technique~\cite{bradley1994} because of its high correlation with psychological response~\cite{lang1993} and focus on the metrics of valence and arousal.
For attitudes, the participants are asked to evaluate the level of overall comfort, naturalness, friendliness, and intelligence using a 7-point Likert questionnaire
(from 1-7, higher scores indicated stronger agreement).
The present study only focused on the initiating process of HRI.
Participants were also instructed to give ratings based on their experience of initiating interaction.

A success case means that the participant use at least one function of the \textit{Xiaodu} robot.
For example, after deploying the proposed framework, the robot may greet then suggest
one function to the participant.
If the participant noticed it and responded, this is a success case.
For the existing wake word-based HRI, the participant should try to wake,
the robot then following text instruction on the screen to make a success interaction case.
A failure case means that the participant neither responds to the proactive calling of the robot nor wakes the robot and follows the instruction.

\begin{figure}
  \centering
  \vspace{0.08in}
  \includegraphics[width=0.8\columnwidth]{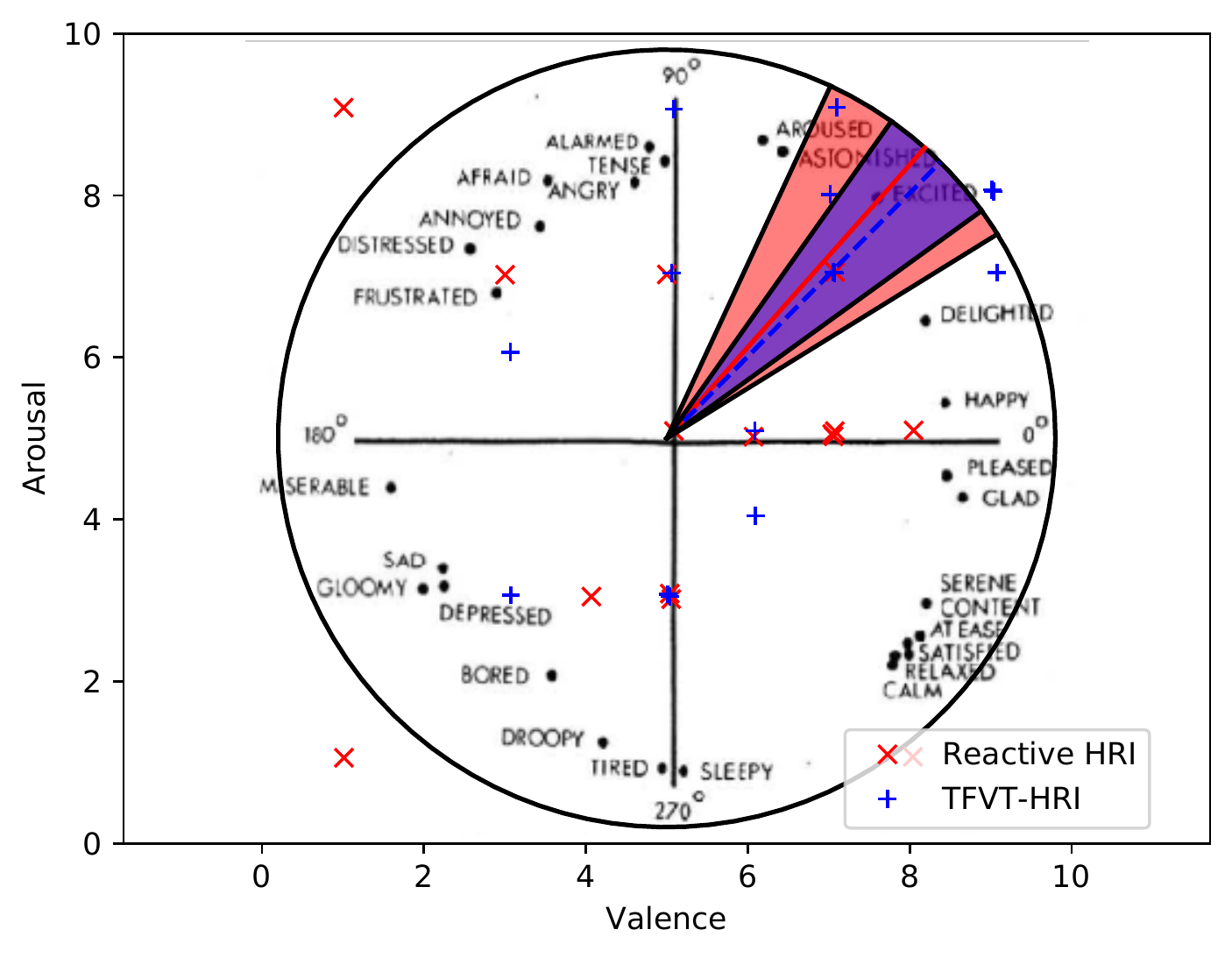}
  \vspace{-0.1in}
  \caption{A scatter diagram of the affects on the circumplex model
    \cite{russell1980circumplex}.
    In the diagram, the blue line and red line are the respective mean of
    the affect (i.e., the overall experience of feeling or emotion) on the proposed TFVT-HRI framework and the reactive HRI system;
    the blue and red sector indicate the variances of our model and passive
    method respectively.
    For checking users' emotions easily, we merge the scatter diagram and
    affect words coordinates together.
    Clearly, both approaches arouse emotions close to EXCITED.}
  \label{fig:va}
  \vspace{-0.15in}
\end{figure}

For the objective metric, success rate, 100\% of participants had success
interactions when using the proposed framework.
Whereas the success rate was 80\% on the existing wake word-based HRI system.
For emotions, the results of descriptive statistics are shown in Table~\ref{tab:user_reports}.
Independent-Samples T Test indicated that there was no significant difference
in valence ($t(28) = 1.218$, $p = 0.233 > 0.05$) and arousal ($t(28) = 1.906$, $p = 0.067 > 0.05$)
We combined the scatter diagram of the valence-arousal plain and affect
words coordinates in Fig.~\ref{fig:va}.
It informed us that two groups both have a positive emotion, EXCITED.
The Levene's Test indicated that all the attitudes variables except for the intelligence score met the assumptions of variance homogeneity at $p > 0.05$.
Thus, we used the Independent-Samples T Test for Overall Comfort Level, Naturalness, Friendliness scores, and T’ Test for intelligence score.
Compared to the control group, the experimental group reported significantly higher scores in Overall Comfort Level ($t(28) = 2.141$, $p = 0.041 < 0.05$), Naturalness ($t(28) = 2.354$, $p = 0.026 < 0.05$), Friendliness ($t(28) = 2.705$, $p = 0.012 < 0.05$), and Intelligence ($t'(24.679) = 2.225$, $p = 0.035 < 0.05$).

In summary, the proposed framework has a higher success rate of initiating an interaction than the existing wake word-based HRI system of the \textit{Xiaodu} robot due to that the proposed framework can capture the intention of the participants and trigger multi-modal interaction signals (language, facial expression, and body language).
However, no significant difference was found in the emotions of the two groups. The robot's characters may have some positive impacts on users' emotions, such as its cute appearance (Fig.~\ref{fig:xiaodu}), and child-like voices. However, the proposed framework leads to significant differences in feelings of overall comfort, naturalness, friendliness, and intelligence.

\subsection{Showcases and the Discovered Social Rules}

\begin{figure}
  \centering
  \vspace{0.08in}
  \includegraphics[width=0.9\columnwidth]{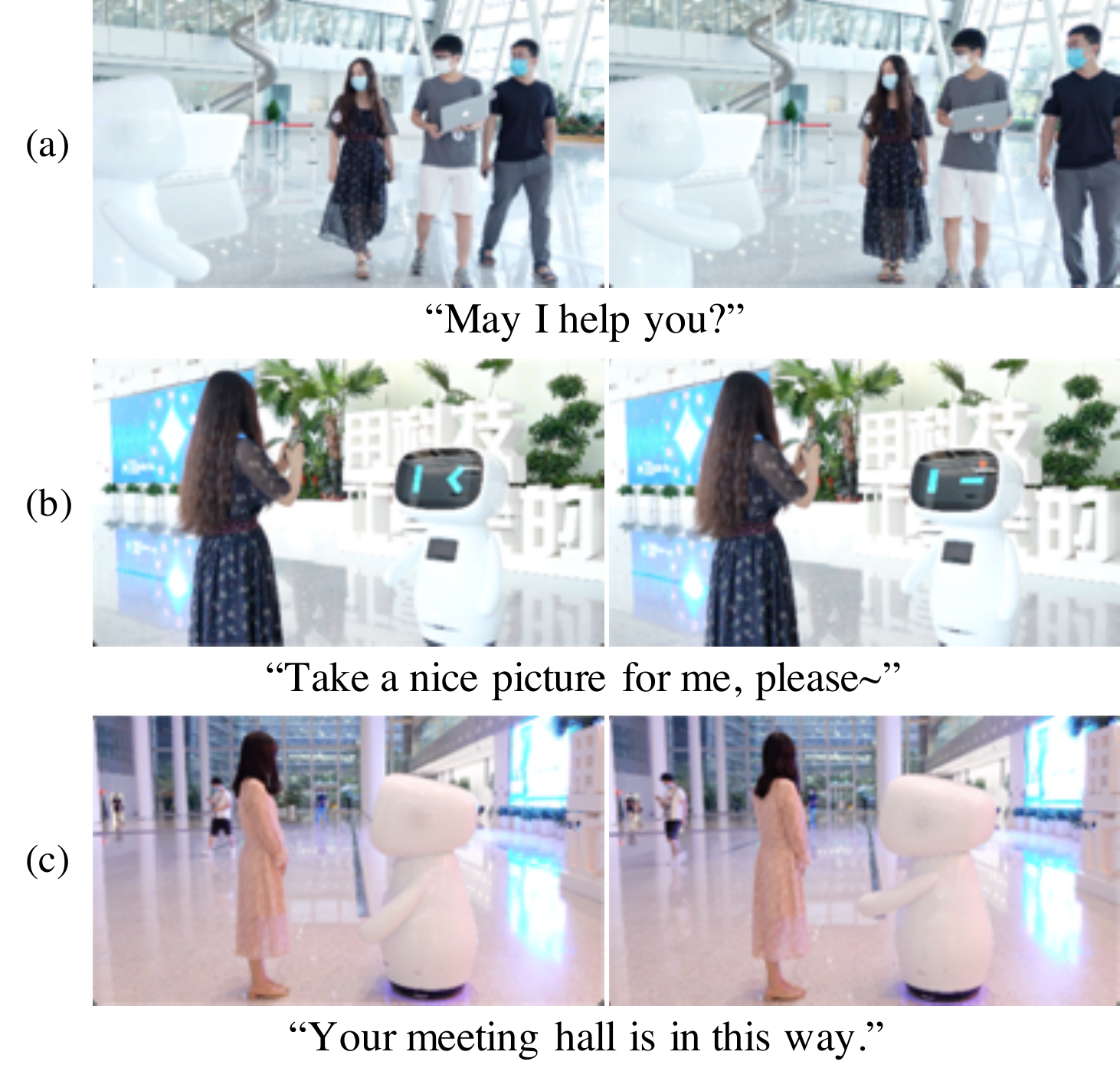}
  \vspace{-0.1in}
  \caption{Showcases that the \textit{Xiaodu} robot interacts with pedestrians following the commonly recognized social rules.}\vspace{-0.2in}
  \label{fig:showcase}
\end{figure}

Some interaction cases are shown in Fig.~\ref{fig:showcase}.
As we can see in Fig.~\ref{fig:showcase} (a), when a group of visitors passed by, the \textit{Xiaodu} robot raised its hand and greeted these people using voice.
Here we emphasize that the \textit{Xiaodu} robot started this action when the visitors were in the near field, matching the prior studies based on Hall's proxemics theory~\cite{hall1966hidden}.
In the second case (Fig.~\ref{fig:showcase} (b)), a visitor were taking photos with the \textit{Xiaodu} robot.
The robot proactively blinked its eyes as responses.
The third case (Fig.~\ref{fig:showcase} (c)) shows that the \textit{Xiaodu} robot successfully led the girl to a depth interaction.
Overall, the TFVT-HRI framework substantially learned certain knowledge of social rules from the experts' demonstrations.

\section{CONCLUSIONS}
In this work, we propose a new end-to-end framework for the proactive HRI task, TFVT-HRI. The proposed framework is featured with object-level modeling, reasoning over time and semantic space, and a relatively large multi-modal action space.
Offline dataset validation and online user study are carried out.
It is proved that the proposed model is better at the HRI task than other end-to-end models such as R(2+1)D, which is also proved to require less computational cost than R(2+1)D. 
The user study has revealed that the proposed framework achieves a substantially higher score in overall comfort, naturalness, friendliness, and intelligence than the existing wake word-based reactive HRI system on the \textit{Xiaodu} robot.
One possible extension of this work is to employ reinforcement learning online utilizing user feedback signals. Also, we could further extend the action space to continuous space to permit the robot to react with a broader range of behaviors, for which it is necessary to explore multi-modal generative models.

\addtolength{\textheight}{-7cm}   








\bibliographystyle{IEEEtran}
\bibliography{IEEEabrv,references}

\end{document}